\documentclass{article}

\usepackage[preprint]{corl_2026} 

\usepackage{graphicx}
\usepackage{amsmath}
\usepackage{amssymb}
\usepackage{mathtools}
\usepackage{amsthm}
\usepackage{multirow}
\usepackage{makecell}
\usepackage{booktabs}
\usepackage{wrapfig}
\usepackage{caption}
\usepackage{subcaption}
\usepackage{float}
\usepackage[normalem]{ulem}

\usepackage[flushmargin]{footmisc}
\newcommand\blfootnote[1]{%
  \begingroup
  \renewcommand\thefootnote{}\footnote{#1}%
  \addtocounter{footnote}{-1}%
  \endgroup
}

\title{DriveMLP: When Training Strategy Outperforms Model Complexity in Autonomous Driving}

\author{
  Alessandro Canevaro$^{\ast, 1, 2}$\\
  \And
  Julian Schmidt$^{1}$\\
  \And
  Hang Yu$^{1, 3}$\\
  \And
  Ye Jin$^{1, 4}$\\
  \AND
  Sajad Marvi$^{1, 5}$\\
  \And
  Georg Martius$^{2}$\\
  \And
  Julian jordan$^{1}$\\
}

\begin{document}
\maketitle


\begin{abstract}
    Learning motion planning policies that remain reliable under closed-loop execution is a central challenge in Autonomous Driving (AD).
    Imitation Learning (IL) scales well to large AD datasets but suffers from distribution shift, causal confusion, and compounding errors under closed-loop execution.
    In contrast, Reinforcement Learning (RL) typically requires handcrafted rule-based rewards and massive simulation.     
    We study a simple closed-loop IL framework that uses RL purely as an optimization tool, guided by a trajectory-distance reward derived from logged expert data.
    We refer to this training strategy as Expert-Guided RL (EGRL).
    Through a controlled empirical study we find evidence that the training procedure, rather than model size and complexity, accounts for a substantial part of the performance gains.
    We then introduce DriveMLP, a lightweight MLP policy trained with EGRL, as a simple instantiation of these findings. It achieves a score of 86.51 on the nuPlan Val14 Reactive benchmark, a +3.2 point improvement over the previous best IL planner (Flow Planner), with $>15\times$ faster inference.
    Code is available at: \newline \url{https://github.com/alessandro-canevaro/DriveMLP}. 
\end{abstract}

\keywords{Autonomous Driving, Closed-Loop Training, Planning} 


\blfootnote{
$^{\ast}${\tt\small alessandro.canevaro@mercedes-benz.com}\\
$^{1}$Mercedes-Benz AG, Research \& Development, Sindelfingen, Germany\\
$^{2}$University of T{\"u}bingen, T{\"u}bingen, Germany\\
$^{3}$Karlsruhe Institute of Technology, ITIV, Karlsruhe, Germany\\
$^{4}$Technical University of Munich, Munich, Germany\\
$^{5}$University of Freiburg, Freiburg, Germany
}

\section{Introduction}

Autonomous Driving (AD) planning requires policies that remain reliable in complex multi-agent interactions and during rare, safety-critical events~\cite{survey}. 
Despite rapid progress in large-scale learning-based planners, a fundamental challenge remains: how to train models that are robust under closed-loop execution without sacrificing efficiency or scalability.

In practice, Imitation Learning (IL) has emerged as the dominant paradigm because it scales well to large open-loop datasets of human driving.
Recent methods improve closed-loop performance through auxiliary objectives and data augmentation, alongside increasingly expressive architectures. 
These include transformer-based~\cite{46201} planners such as PlanTF~\cite{cheng2023plantf} and PLUTO~\cite{cheng2024pluto}, as well as generative approaches such as Diffusion Planner~\cite{zheng2025diffusionbased} and Flow Planner~\cite{tan2025flow}.

However, despite the advances, IL fundamentally suffers from \emph{distribution shift} and \emph{causal confusion}: models are trained in an open-loop setting, where spurious correlations in the data may be learned, and are then deployed in closed-loop, where compounding errors can push the policy into previously unseen states~\cite{de2019causal,imitationisnotenough}. 
A telling example of causal confusion is that Diffusion Planner loses nearly 10 percentage points in closed-loop performance when provided with additional ego-state inputs~\cite{zheng2025diffusionbased}, suggesting that open-loop training can turn useful causal cues into spurious shortcuts. 
More broadly, under mild test-time distribution shifts, such as the nuPlan reactive benchmark, where surrounding agents are controlled by rule-based IDM~\cite{idm_article} policies instead of replayed logs, open-loop IL methods exhibit substantial performance degradation. 
These observations motivate training objectives that expose the policy to its own induced state distribution.

Reinforcement Learning (RL) offers this closed-loop training signal, but applying it to AD remains challenging~\cite{misdesign, 2022NaturGranTurismo}. 
Strong results often depend either on heavily engineered rule-based reward functions, as in CaRL~\cite{jaeger2025carl}, or on simpler, though still rule-based, rewards paired with extremely large-scale simulation, as in GIGAFLOW~\cite{cusumano-towner2025robust}.

A natural alternative is to derive supervision from expert demonstrations rather than hand-designed rules.
Online Inverse Reinforcement Learning (IRL) methods such as GAIL~\cite{GAIL2016Ermon} and AIRL~\cite{fu2018learning} learn rewards via adversarial training, while non-adversarial alternatives such as IQ-Learn~\cite{NEURIPS2021_210f760a} rely on occupancy matching.
However, both families are computationally expensive and difficult to stabilize at scale~\cite{zhao2025beyondexpert}.
Recent hybrid IL-RL approaches reduce or eliminate adversarial reward learning by analytically deriving demonstration-based rewards, but  introduce other forms of complexity.
CarPlanner~\cite{carplanner} uses RL as a trajectory candidate generator, requiring a secondary rule-based post-processing module for final selection and refinement.
Peng et al.~\cite{peng2024improving} leverage expert-guided trajectory signals for multi-agent traffic simulation, employing a two-stage training strategy where RL fine-tunes a pre-trained autoregressive policy.
Another approach is BC-SAC~\cite{imitationisnotenough}, which alternates BC and RL updates using safety-related rewards, treating RL primarily as a corrective component.

In summary, recent ego-planning methods rely on demonstration-based supervision using complex architectures that sacrifice inference speed or, orthogonally, on engineered rule-based supervision.

These approaches motivate an empirical question: how competitive can a simple planning policy become when trained in closed-loop simulation with demonstration-derived supervision?
We investigate a practical training recipe and evaluate its performance against SOTA methods on Highway and nuPlan benchmarks.
We further assess whether this recipe enables a lightweight MLP to compete with recent generative imitation-based planners.

Additional related work and references are discussed in Appendix~\ref{app:extended_related_work}.

\begin{wrapfigure}{r}{0.5\textwidth}
\vspace{-15pt}
    \begin{center}
        \includegraphics[width=1.0\linewidth]{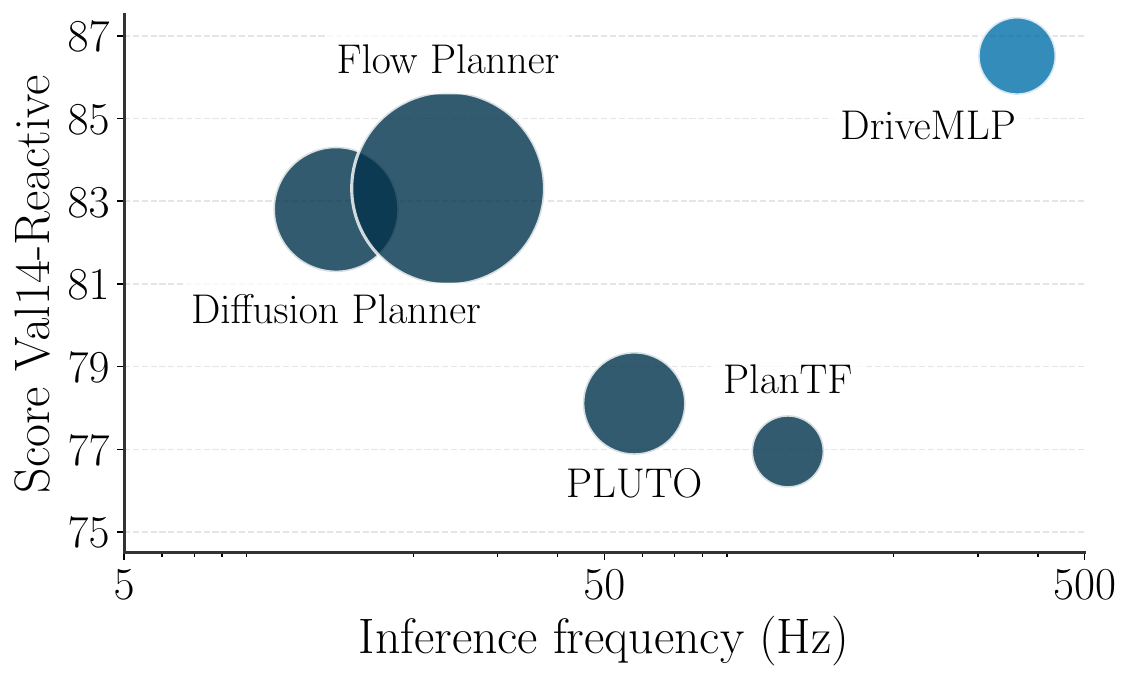}
    \end{center}
    \caption{
    DriveMLP combines strong reactive performance with low inference cost.
    The plot shows nuPlan Val14-Reactive scores against inference frequency, with circle size indicating parameter count.
    DriveMLP, a lightweight MLP trained with EGRL, achieves higher scores and faster inference than the compared SOTA planners.
    }
    \label{fig:intro_radar}
\end{wrapfigure}

In this work, we study closed-loop imitation through DriveMLP, a lightweight MLP planner trained using standard RL in a non-reactive log-replay simulator.
Training is guided by a reward derived from trajectory deviations from the logged expert, a strategy we name Expert-Guided RL (EGRL).
Our main contributions are as follows: 
\textbf{(1)}~Through a controlled empirical analysis, we show that performance gains stem largely from the training procedure, rather than increased model size or complexity alone.
\textbf{(2)}~We introduce DriveMLP, a lightweight MLP policy trained with EGRL, as a simple instantiation of these findings. 
On the evaluated nuPlan reactive splits, it achieves SOTA performance among purely learned imitation-based planners while running $>15\times$ faster than recent generative planners.
\textbf{(3)}~We propose an exponential reward formulation for AD building on~\cite{expreward} that provides dense and stable supervision, improving over the Peng et al.~\cite{peng2024improving} formulation.
\textbf{(4)}~We show that through closed-loop RL exploration, our model is substantially more robust to action noise and data efficient compared to SOTA open-loop IL.

\section{Method}
This section describes DriveMLP and its EGRL training procedure.
We first formalize the problem and non-reactive simulation setup, then introduce the trajectory-distance reward and the policy architecture.

\subsection{Problem Formulation}

We formulate AD planning as a Markov Decision Process defined by the tuple $(\mathcal{S}, \mathcal{A}, \mathcal{T}, \mathcal{R}, \gamma)$, where $\mathcal{S}$ is the state space, $\mathcal{A}$ is the action space, $\mathcal{T}: \mathcal{S} \times \mathcal{A} \times \mathcal{S} \rightarrow [0,1]$ is the transition function, $\mathcal{R}: \mathcal{S} \times \mathcal{A} \rightarrow \mathbb{R}$ is the reward function, and $\gamma \in [0,1]$ is the discount factor.

\textbf{State Space.} The state $s_t \in \mathcal{S}$ consists of: 
(1) vectorized map representation including lane geometry, traffic light status, and route information, 
(2) surrounding dynamic agents with their positions, velocities, and headings, 
and (3) static objects in the environment. 

Unlike most IL models, we additionally provide the agent with ego-vehicle dynamic state information (velocity, acceleration, steering angle, and yaw rate).
All spatial information is represented in the ego-centric coordinate frame.

\textbf{Action Space.} We adopt the low-level control formulation, where actions $\mathbf{a}_t \in \mathcal{A}$ consist of normalized steering angle and acceleration commands: $\mathbf{a}_t = (\phi_t, a_{t}) \in [-1, 1]^2$. 
The actions are then scaled to the vehicle model's steering and acceleration ranges by the simulator itself.

\textbf{Expert Demonstrations.} We assume access to a dataset $\mathcal{D} = \{\tau^i\}_{i=1}^N$ of $N$ expert driving state trajectories, where each trajectory $\tau^i = (s_0^i, \ldots, s_{T_i}^i)$.

\subsection{Non-Reactive Simulation}

\begin{figure*}[ht]
  \begin{center}
    \centerline{\includegraphics[width=\linewidth]{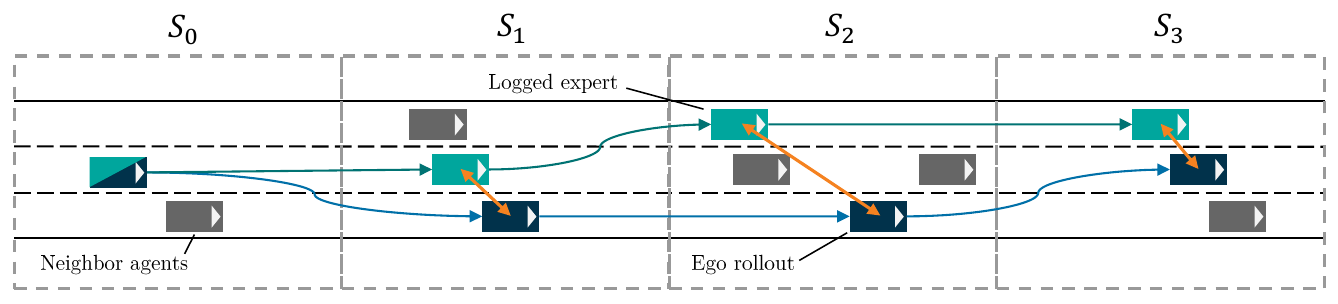}}
    \caption{
      Working principle of the non-reactive simulator. The teal rectangles represent the logged expert ego trajectory. 
      The gray rectangles represent other agents or obstacles moving according to recorded data, while the dark blue rectangle shows the ego rollout following the actions of the learned policy. 
      Orange arrows show the position distance between the ground truth and the ego rollout which is then used in the reward computation.
    }
    \label{fig:nr_sim}
  \end{center}
\vspace{-15pt}
\end{figure*}

A key design principle of our approach is training in a \emph{non-reactive} simulation environment. 
Unlike standard RL environments where all agents respond dynamically to the ego vehicle's actions, our environment operates as follows (Figure~\ref{fig:nr_sim}):

\textbf{Ego vehicle:} Controlled by the learning agent, following the dynamic model constraints with actions sampled from the policy $\pi_\theta$.\\
\textbf{Other agents:} Follow their pre-recorded trajectories from the dataset, independently of the ego vehicle's behavior.\\
\textbf{Static environment:} Road geometry, traffic lights, and static obstacles remain fixed as in the original recorded scenario.

Our reward formulation (Section \ref{sec:reward_func}) is not applicable in fully reactive environments as it would not be well-defined.
Deviations by the ego vehicle would causally alter the behavior of other agents, invalidating the logged expert trajectory as a reference and leading to ambiguous or misleading supervision. 
By replaying surrounding agents from logs, the expert ego trajectory remains a consistent reference target, enabling dense, step-wise supervision without adversarial reward learning or expert policy access.

\subsection{Trajectory-Based Reward Function}
\label{sec:reward_func}
Given access to expert trajectories, we define a simple reward function that measures the agent's deviation from expert behavior based on position and speed.
Unlike rule-based RL rewards that encode task-specific driving heuristics, our reward directly reflects expert behavior. 
The only additional constraints that we impose are termination conditions that define the domain in which replay-based imitation remains valid, rather than preferences over driving behavior.

Minimizing trajectory deviations can be interpreted as maximizing the likelihood of expert state visitation under a Gaussian noise model on system dynamics. 
In this sense, our reward implements a closed-loop analog of BC, where supervision is applied to realized state trajectories, induced by the policy itself, rather than one-step expert actions.
As a result, errors compound differently and the policy is explicitly trained to recover from its own deviations.
Appendix~\ref{app:theory} provides a deeper discussion on closed-loop regret bounds.

At each timestep $t$, let $p_t^{\text{ego}} \in \mathbb{R}^2$ denote the ego vehicle's position and $p_t^{\text{expert}} \in \mathbb{R}^2$ denote the corresponding expert position. 
We define the distance between ego and expert at timestep $t$ as:
\begin{equation}
\delta_t = \|p_t^{\text{ego}} - p_t^{\text{expert}}\|_2
\end{equation}

We then map the normalized distance to $[0,1]$ using an exponential function:
\begin{equation}
r_t^{pos} = \exp\left( -\alpha_1\frac{\delta_t}{\kappa_t^{\text{expert}} + \epsilon}\right), \quad \text{with} \quad \kappa_t^{\text{expert}} = \|p_t^{\text{expert}} - p_{t-1}^{\text{expert}}\|_2,
\end{equation}

where $\epsilon$ is a small constant that prevents division by zero when the expert vehicle is stationary.
We use the expert's displacement $\kappa_t^{\text{expert}}$ as a normalization factor to account for speed-dependent tolerance, since higher speeds naturally lead to larger positional deviations for the same control error.
Similarly, velocity deviations between the ego and the expert are penalized via:
\begin{equation}
r_t^{vel} = \exp \left( -\alpha_2\|v_t^{\text{ego}} - v_t^{\text{expert}}\|_2\right),
\end{equation}

where $v_t \in \mathbb{R}$ denotes scalar longitudinal speed (m/s).
The full reward expression is then:
\begin{equation}
r_t = \alpha_3 r_t^{pos} + (1-\alpha_3) r_t^{vel}
\end{equation}

where $\alpha_{1, 2, 3}$ are hyperparameters that regulate the scale of the different reward terms.

\textbf{Episode Terminations.} The trajectory-distance reward is meaningful only within a neighborhood of the expert trajectory, where interactions between the ego and other agents remain approximately consistent with the logged data.
We enforce termination conditions that define the domain in which the imitation reward is valid. 
When the ego vehicle deviates beyond this domain, the episode is terminated and assigned a fixed penalty $r_p$.
Specifically, the termination criteria are:
(1) the ego vehicle goes outside the drivable area, 
(2) the ego vehicle bounding box overlaps with the ones of other agents or obstacles, 
or (3) the ego vehicle deviates significantly from the expert trajectory.
Further details are reported in the Appendix~\ref{app:hyperparams}.

\subsection{Policy Learning}
The policy network $\pi_\theta: \mathcal{S} \rightarrow \Delta(\mathcal{A})$ and the value function $V_\phi: \mathcal{S} \rightarrow \mathbb{R}$ are parameterized as separate multi-layer perceptrons (MLPs).

To encode variable-length inputs (e.g., neighboring agents and map polylines), we use permutation-invariant set encoders~\cite{NIPS2017_f22e4747}. 
Specifically, we employ dedicated encoders for the ego state, map, route, neighboring agents, and static obstacles. 
Each encoder is composed of a shared MLP applied to each input element, followed by mean pooling to obtain a fixed-dimensional representation and a second MLP to further process the aggregated latent features.
The resulting encoder outputs are concatenated and mapped to a joint latent representation using an additional MLP, which is then fed to the policy and value heads. 
The policy head parameterizes a diagonal Gaussian distribution over the continuous action space, predicting both the mean and (diagonal) variance.

The model's parameters can be optimized with standard RL algorithms.
As later shown in the Experiments section, we tested two established ones: Soft Actor-Critic (SAC)~\cite{haarnoja2017soft} and Proximal Policy Optimization (PPO)~\cite{schulman2017ppo}.

\section{Experiments}
We evaluate DriveMLP on two benchmarks to assess whether our method improves closed-loop performance and robustness. 
We describe the experimental setup and then report quantitative results.

\subsection{Experimental Setup}

We distinguish between three evaluation settings. 
First, in controlled comparisons (Section~\ref{sec:ctrl_comp}) we train identical models on the same data with different learning objectives, isolating the effect of training strategy itself. 
Second, we apply our training strategy to recent SOTA models to assess its applicability to different architectures (Section~\ref{sec:ctrl_comp2}).
Lastly, we evaluate whether a lightweight model trained via our closed-loop optimization is sufficient to outperform SOTA nuPlan IL baselines (Section~\ref{sec:sota_comp}). 
To ensure a fair comparison of learning paradigms, we focus our analysis on methods trained primarily from human demonstrations and leave out rule-based and RL planners that mainly optimize traffic-rule-based reward or cost functions (e.g., speed limits, lane keeping, comfort, or progress terms).

\textbf{Benchmarks.} We evaluate on two AD benchmarks of increasing complexity: 
(1) \textbf{Highway-env}~\cite{highway-env}: A lightweight multi-lane highway driving simulator featuring dense traffic and lane changes. We use the default configuration for observations and the provided continuous control action space. (2) \textbf{nuPlan}~\cite{nuplan,karnchanachari2024nuplan}:  A large-scale, closed-loop planning benchmark built on 1,500+ hours of real-world human driving logs collected across 4 cities (Las Vegas, Pittsburgh, Boston, and Singapore), making it one of the strongest simulation proxies for real-world urban driving currently available~\cite{karnchanachari2024nuplan,jaeger2025carl}.
We train on scenarios from the training split and evaluate on three standard test sets: Val14~\cite{Dauner2023CORL}, Test14-Hard~\cite{cheng2023plantf}, and Test14~\cite{cheng2023plantf}.     
Additional results on the InterPlan~\cite{10803052} benchmark are available in Appendix~\ref{app:interplan}.

\textbf{Implementation and evaluation.}
To isolate the effectiveness of our closed-loop RL training with trajectory-based rewards, we intentionally avoid common training augmentations in our method, including auxiliary losses, curriculum learning, multi-source training data, and advanced model architectures.
For nuPlan, we follow the official closed-loop evaluation protocol~\cite{nuplan} and report the overall score under both Reactive (R) and Non-Reactive (NR) settings.
Full implementation details, hyperparameters, dataset preparation, and evaluation metrics are provided in Appendix~\ref{app:setup}.

\subsection{Controlled Training Strategy Comparison}
\label{sec:ctrl_comp}

\textbf{Baselines.} We compare our training strategy against BC, AIRL~\cite{fu2018learning}, and BC-RL~\cite{imitationisnotenough} while using the same network architecture for all methods. 
For all training strategies except BC, we use SAC as the backbone RL optimizer on Highway-Env and PPO on nuPlan.

\textbf{Results.} Table~\ref{tab:combined_results} reports performance on both benchmarks. 
On Highway-Env, our method substantially outperforms BC across all metrics and achieves performance comparable to AIRL and BC-RL. 
AIRL takes three times more training steps to converge compared to our approach, and each iteration is further slowed down by the reward learning.

On the large-scale nuPlan benchmark, the differences between training strategies become even more pronounced. 
In particular, AIRL remains the most challenging method to scale due to its intrinsic instability. 
To mitigate this, we incorporated discriminator regularization techniques commonly used in GAN training, specifically one-sided label smoothing~\cite{gan_label} and gradient penalty~\cite{gan_grad}.

\begin{table}[ht]
\centering
\caption{Controlled training strategy comparison on Highway-Env (left) and nuPlan (right). 
Highway-Env results averaged over 3 seeds. NR = Non-Reactive, R = Reactive.
All training methods use the same DriveMLP policy architecture.}
\label{tab:combined_results}
\resizebox{0.95\columnwidth}{!}{
\begin{tabular}{lccc|cccccc}
\toprule
& \multicolumn{3}{c|}{Highway-env} 
& \multicolumn{2}{c}{Val14} 
& \multicolumn{2}{c}{Test14-Hard} 
& \multicolumn{2}{c}{Test14} \\
\cmidrule(lr){2-4} \cmidrule(lr){5-6} \cmidrule(lr){7-8} \cmidrule(lr){9-10}
Method 
& Reward $\uparrow$ & Ep. Len $\uparrow$ & Term. $\downarrow$ 
& NR & R & NR & R & NR & R \\
\midrule
BC 
& 17.20 {\scriptsize $\pm$ 3.26} & 23.56 {\scriptsize $\pm$ 4.28} & 48.7\% {\scriptsize $\pm$ 21.48} 
& 53.64 & 32.36 & 17.06 & 19.50 & 30.36 & 33.01 \\

AIRL 
& 20.37 {\scriptsize $\pm$ 0.40} & 28.48 {\scriptsize $\pm$ 0.46} & 9.3\% {\scriptsize $\pm$ 2.87}
& 51.74 & 58.74 & 43.00 & 52.09 & 48.88 & 61.15 \\

BC-RL
& 20.28 {\scriptsize $\pm$ 0.92} & 28.70 {\scriptsize $\pm$ 1.15} & 20.0\% {\scriptsize $\pm$ 17.38} 
& 77.19 & 78.05 & 66.05 & 65.35 & 77.17 & 77.23 \\

\midrule
\textbf{EGRL} & \textbf{21.02 {\scriptsize $\pm$ 0.79}} & \textbf{29.17 {\scriptsize $\pm$ 1.01}} & \textbf{7.3\% {\scriptsize $\pm$ 3.55}}
& \textbf{90.38} & \textbf{86.51} & \textbf{76.68} & \textbf{74.50} & \textbf{86.38} & \textbf{87.06} \\
\bottomrule
\end{tabular}
}
\end{table}

\subsection{Transfer Across Planner Backbones}
\label{sec:ctrl_comp2}

To assess whether the benefits extend beyond MLP policies, we use EGRL to train policies based on feature encoders adapted from PlanTF~\cite{cheng2023plantf}, PLUTO~\cite{cheng2024pluto}, and Diffusion Planner~\cite{zheng2025diffusionbased}.

We adapt these models for closed-loop RL training by (1) modifying the input representation to include ego-state features,
(2) adjusting the output space from multi-step trajectory prediction to single-step control actions, 
and (3) removing open-loop-specific auxiliary heads and objectives such as joint multi-agent prediction.
These adaptations are necessary to interface existing SOTA IL planners with our on-policy closed-loop training framework while preserving the underlying transformer backbone.
Extending to iterative generative heads remains an interesting direction for future work as it requires online RL through computationally intensive multi-step denoising loops.

Table~\ref{tab:drivemlp_sota_backbones} shows that applying the proposed training objective to encoders derived from existing planners consistently improves reactive performance, matching or exceeding the original pipelines.

\begin{table}[h]
\centering
\caption{EGRL training strategy applied to other SOTA backbones.
*Flow Planner uses a similar encoder structure as Diffusion Planner.
}
\label{tab:drivemlp_sota_backbones}
\resizebox{0.6\columnwidth}{!}{
\begin{tabular}{lcccc}
\toprule
Model & Original (Val14-R) & \textbf{+EGRL} & $\Delta \%$ & Params \\
\midrule
PlanTF & 76.95 & 78.26 & +1.70 & 2.0M \\
PLUTO w/o ref. & 78.11 & 83.41 & +6.78 & 2.4M \\
Diff. Planner* & 82.80 & 86.05 & +3.92 & 2.3M \\
\bottomrule
\end{tabular}
}
\end{table}

Additional experiments across MLP capacities and with a reduced transformer policy are reported in Appendix~\ref{app:agnosticism}.

\subsection{Comparison with State of the Art}
\label{sec:sota_comp}

\textbf{Baselines.} We evaluate against SOTA IL approaches, including PlanTF~\cite{cheng2023plantf}, PLUTO~\cite{cheng2024pluto}, and the more recently proposed diffusion-based~\cite{zheng2025diffusionbased} and flow-matching-based~\cite{tan2025flow} planners that incorporate generative modeling architectures.

\textbf{Results.} In particular, our training strategy enables a simple MLP model with approximately 2.3M parameters to outperform several SOTA IL baselines in the reactive settings, as shown in Table~\ref{tab:nuplan_results}.
Our approach exhibits a significantly smaller performance gap between non-reactive and reactive simulation settings, indicating improved robustness and generalization to interactive environments.

We hypothesize that open-loop IL methods overfit to the non-reactive training distribution: by learning to imitate the expert conditioned on the observed agent configurations, these methods may implicitly rely on correlations between agent states and correct actions that hold when agents follow their logged trajectories, but break when agents react to the ego instead.
DriveMLP is also trained in a non-reactive simulator, but its closed-loop objective may encourage the policy to rely less on such scene-level correlations and more on ego dynamics.
This interpretation is consistent with ego-state ablation (Appendix~\ref{app:ego_state}), which shows that ego-state features help DriveMLP but hurt open-loop IL methods, suggesting that closed-loop training helps establish more correct causal relationships between ego dynamics and planning decisions.

\begin{table}[ht]
\centering
\caption{Overall scores on nuPlan benchmark. NR = Non-Reactive, R = Reactive simulation.}
\label{tab:nuplan_results}
\resizebox{0.62\columnwidth}{!}{
\begin{tabular}{lcccccc}
\toprule
& \multicolumn{2}{c}{Val14} & \multicolumn{2}{c}{Test14-Hard} & \multicolumn{2}{c}{Test14} \\
\cmidrule(lr){2-3} \cmidrule(lr){4-5} \cmidrule(lr){6-7}
Method & NR & R & NR & R & NR & R \\
\midrule
PlanTF & 84.27 & 76.95 & 69.70 & 61.61 & 85.62 & 79.58 \\
PLUTO w/o refine & 88.89 & 78.11 & 70.03 & 59.74 & \textbf{89.90} & 78.62 \\
Diffusion planner & 89.87 & 82.80 & 75.99 & 69.22 & 89.19 & 82.93 \\
Flow Planner & \textbf{90.43} & 83.31 & 76.47 & 70.42 & 89.88 & 82.93 \\
\midrule
\textbf{DriveMLP} & 90.38 & \textbf{86.51} & \textbf{76.68} & \textbf{74.50} & 86.38 & \textbf{87.06} \\
\bottomrule
\end{tabular}
}
\end{table}

\section{Ablations and Further Analysis}
We analyze DriveMLP's robustness and efficiency through targeted ablations on reward formulation, training dynamics curves, action noise, data efficiency with reduced demonstrations, and inference latency.
Additional parameter sensitivity analysis, including reward ablations, action frequency, and ego-state effects is provided in Appendix~\ref{app:param_sensitivity}.

\subsection{Reward Formulation Ablation}
\label{sec:reward_ablation}

\begin{wraptable}{r}{0.35\columnwidth}
\centering
\vspace{-13pt}
\caption{Exponential reward ablation: NuPlan Val14.}
\label{tab:reward}
\resizebox{0.35\columnwidth}{!}{
\begin{tabular}{lcc}
\toprule
Reward & NR & R \\
\midrule
\textbf{Base} &  \textbf{90.38} & \textbf{86.51} \\
\midrule
w/o speed reward   & 87.44 & 84.92 \\
w/o exp. & 84.87 &  82.74 \\
w/o normalization & 50.49 & 47.58 \\
\bottomrule
\end{tabular}
}
\vspace{-5pt}
\end{wraptable}

Table~\ref{tab:reward} summarizes the impact of removing individual components of our reward formulation. 
We ablate (1) the speed term $r_t^{vel}$, (2) the exponential normalization by replacing it with the inverse distance, and (3) the normalization by directly using the raw distance $\delta_t$, following Peng et al.~\cite{peng2024improving}. 
The results show that each component contributes to performance, with normalization being particularly important for stable and effective training.

\subsection{Training Dynamics}
\label{app:training_dynamics}

\begin{figure}[hbt]
  \centering

  \begin{subfigure}{0.43\columnwidth}
    \centering
    \includegraphics[width=\linewidth]{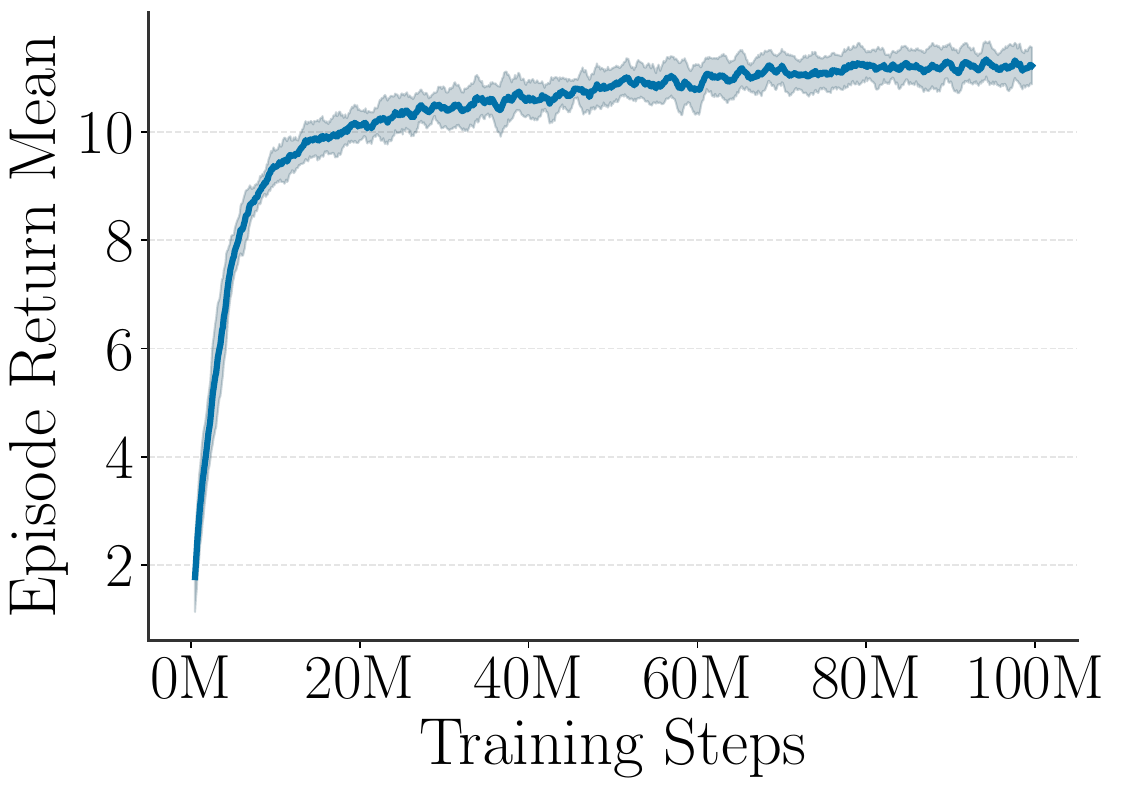}
  \end{subfigure}
  \hfill
  \begin{subfigure}{0.43\columnwidth}
    \centering
    \includegraphics[width=\linewidth]{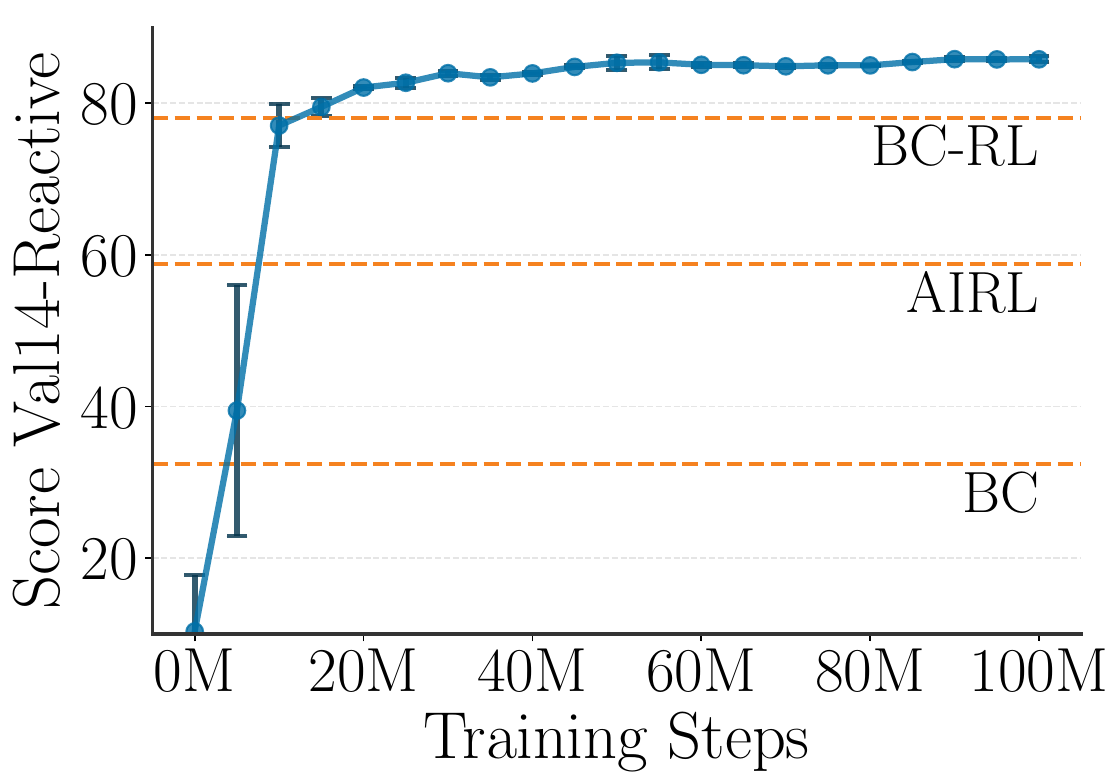}
  \end{subfigure}

  \caption{
    Training dynamics and validation performance over training steps. 
    Left: Episode return mean. 
    Right: Val14 R Score. 
    Horizontal dashed lines indicate the BC, AIRL, and BC-RL baselines.
  }
  \label{fig:training_combined}

\end{figure}
Figure~\ref{fig:training_combined} illustrates the training dynamics of DriveMLP trained using PPO on the nuPlan benchmark. 
The agent achieves rapid initial learning within the first 20M steps, surpassing the BC baseline early in training. 
The training reward continues to improve until plateauing around 80M steps, while the validation performance plateaus earlier at approximately 50M steps, ultimately surpassing the SOTA baseline.

\subsection{Mitigating Distribution Shifts}
\label{sec:noise}

To study the robustness benefits of closed-loop training, we evaluate DriveMLP and Diffusion Planner on nuPlan Val14 while injecting Gaussian noise into the actions passed to the vehicle dynamics model.

\begin{wrapfigure}{r}{0.6\columnwidth} 
\centering
\vspace{-20pt}
\includegraphics[width=0.6\columnwidth]{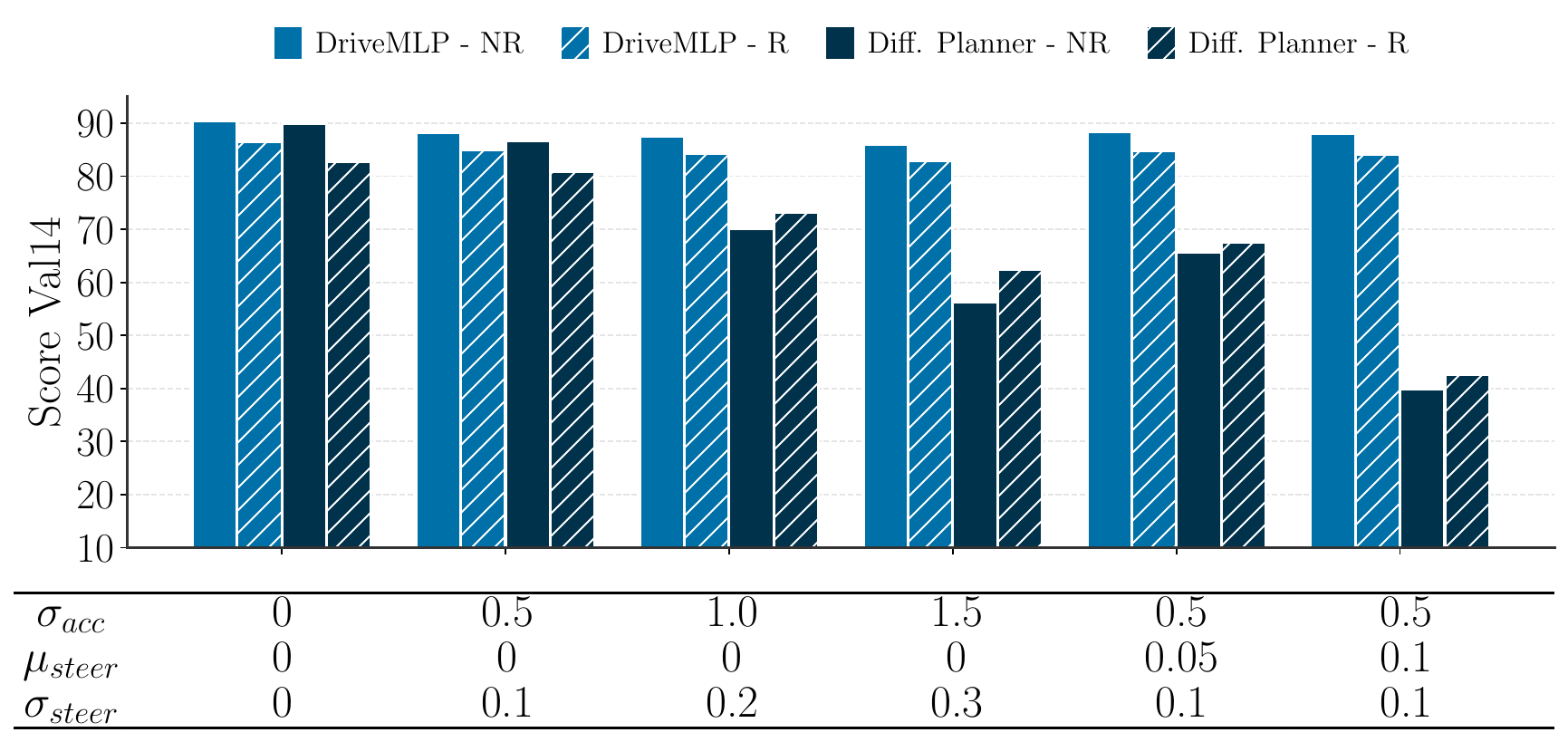}
\caption{
  NuPlan score on Val14 Non-Reactive (solid bars) and Reactive (striped bars) benchmark for DriveMLP (light blue) and Diffusion Planner (dark blue) with Gaussian noise on acceleration ($\sigma_{\text{acc}}$) and on steering ($\mu_{\text{steer}}$, $\sigma_{\text{steer}}$).
}
\label{fig:noise}
\vspace{-15pt}
\end{wrapfigure}

While this form of perturbation does not represent a general semantic distribution shift, it captures a critical and practically relevant failure mode in AD, where small execution errors can accumulate over time and lead to substantial trajectory drift.

Figure~\ref{fig:noise} summarizes performance under varying noise magnitudes. 
Although Diffusion Planner already leverages data augmentation during training to improve robustness to perturbations, DriveMLP exhibits a substantially smaller degradation in closed-loop performance across all noise settings.
Appendix \ref{app:noise} provides a visual comparison of representative trajectories.

\subsection{Data Efficiency Analysis}
\label{sec:data_eff}

\begin{wrapfigure}{r}{0.52\columnwidth} 
\centering
\vspace{-15pt}
\includegraphics[width=0.52\columnwidth]{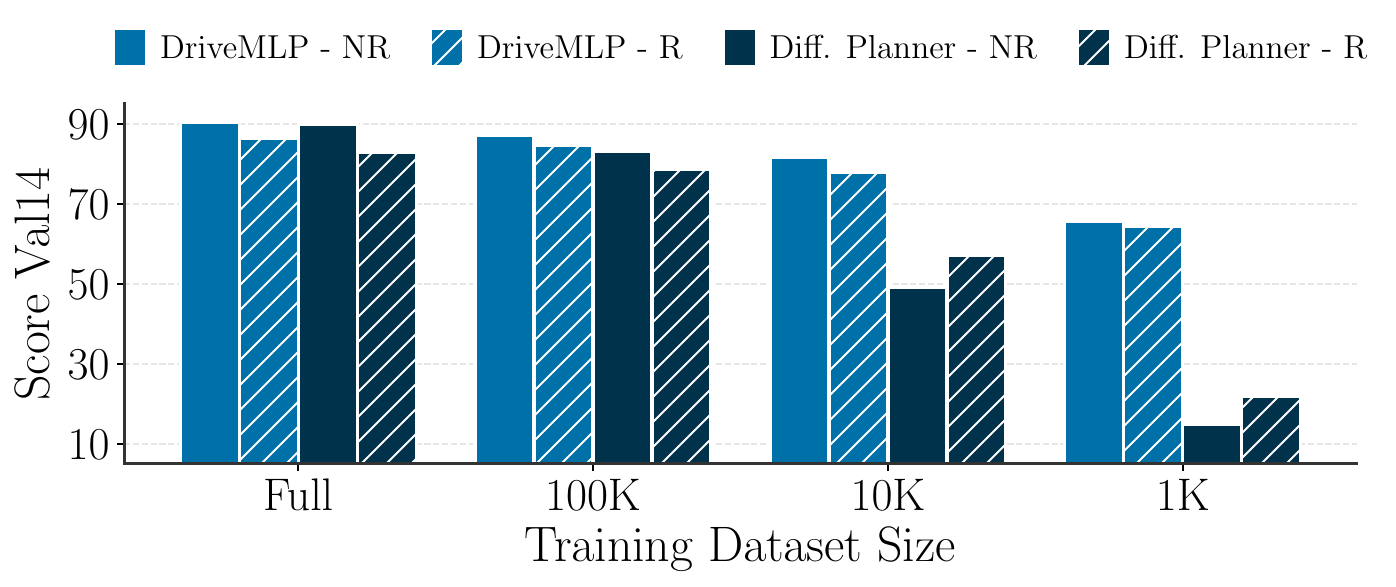}
\caption{
  NuPlan score on Val14 Non-Reactive (solid bars) and Reactive (striped bars) benchmark for DriveMLP (light blue) and Diffusion Planner (dark blue) with reduced training dataset sizes.
}
\label{fig:data_eff_plot}
\vspace{-10pt}
\end{wrapfigure}

A key hypothesis of our approach is that RL-based exploration can compensate for limited expert demonstrations.
Figure~\ref{fig:data_eff_plot} illustrates performance as a function of training dataset size on nuPlan.

DriveMLP maintains strong performance even when trained on significantly reduced demonstration sets, indicating its ability to generate informative training data through exploration.
This property is particularly valuable for long-tail and rare driving scenarios where expert data is scarce.

\subsection{Inference and Training Cost}

\begin{wraptable}{r}{0.55\columnwidth} 
\centering
\caption{Inference speed (NVIDIA P6000), model size, and nuPlan Val14-R performance.}
\label{tab:inference}
\resizebox{0.55\columnwidth}{!}{
\begin{tabular}{lccc}
\toprule
Method & Time (ms) $\downarrow$ & Params (M) $\downarrow$ & Val14-R $\uparrow$ \\
\midrule
PlanTF  &  8.30       & \textbf{2.02}      & 76.95 \\
PLUTO w/o refine   & 17.33        & 4.06      & 78.11\\
Diffusion Planner       & 72.34        & 6.04      & 82.80 \\
Flow Planner          & 42.31        & 14.27      & 83.31 \\
\midrule
\textbf{DriveMLP}       & \textbf{2.76} & 2.33 & \textbf{86.51} \\
\bottomrule
\end{tabular}
}
\vspace{-10pt}
\end{wraptable}

Table~\ref{tab:inference} compares inference latency across methods.
Thanks to its lightweight architecture, DriveMLP is significantly faster than all other SOTA methods, achieving $>15\times$ speedup over Flow Planner while delivering higher Val14-R performance.

In terms of training, DriveMLP introduces additional data-collection overhead arising from closed-loop RL rollouts. 
It requires approximately 112 NVIDIA L4 GPU hours, $\sim$4.7$\times$ BC.
For reference, large-scale IL models such as Diffusion Planner require approximately 240 NVIDIA A100 GPU-hours in our reproduced experiments.
Normalizing by TFLOPS, DriveMLP achieves better NuPlan reactive performance with $\sim$3-5$\times$ less compute than Diffusion Planner.
Full hardware and software details including simulator throughput are provided in Appendix~\ref{app:reproducibility}.

\section{Limitations}
Like all demonstration-guided methods, the trajectory-distance reward anchors learning to the quality of the expert demonstrations.
Nonetheless, our data-efficiency ablation (Section~\ref{sec:data_eff}) shows that closed-loop RL exploration partially compensates for scarce demonstrations, maintaining strong performance even with substantially reduced datasets.
Moreover, EGRL training requires access to a non-reactive log-replay simulator.
While simpler than a fully reactive environment, such a simulator still requires a dynamic vehicle model for the ego.

Although nuPlan provides a high-fidelity closed-loop benchmark grounded in over 1,500 hours of real-world driving data~\cite{karnchanachari2024nuplan}, Sim2Real transfer for object-level planning policies remains challenging. 
While Sim2Real has been successfully demonstrated in other robotics domains~\cite{kaufmann2023champion,miki2022learning,lin2025sim2real}, as well as for sensor-based driving systems~\cite{mueller2018driving}, bridging this gap for object-level planners in AD has yet to be fully realized~\cite{jaeger2025carl}. 
Several promising directions have been explored in prior work~\cite{hanselmann2025emperror,chen2019lbc,jia2023driveadapter}.
Given DriveMLP's low inference latency and robustness, we see integrating it into such pipelines as a natural next step toward real-world deployment.

Appendix~\ref{app:failure} illustrates representative failure cases including overshooting turns, late braking, and missed gap acceptance, providing further insight into the method's current limitations.

\section{Conclusion}
We presented a simple yet effective approach for AD planning that combines the data efficiency of IL with the robustness of RL. 
By training RL agents in non-reactive simulation with trajectory-based rewards, we reduce the distribution shift problems of BC and the reward engineering burden of standard RL.

Our experiments demonstrate that EGRL (1) outperforms other training strategies such as BC, AIRL and BC-RL, (2) improves reactive performance of other SOTA models, and (3) DriveMLP scales to the challenging nuPlan benchmark with SOTA performance on the reactive evaluation while using a simpler architecture and achieving $>15\times$ faster inference compared to the previous best IL model.

Additionally, DriveMLP exhibits superior data efficiency compared to the SOTA Diffusion Planner, maintaining strong performance even with significantly reduced demonstration sets.
By inherently exposing the model to execution-time deviations during training, our method effectively mitigates the accumulation of small execution errors that typically lead to degraded performance of open-loop trained models in closed-loop settings. 

We envision DriveMLP as an easy-to-implement, competitive baseline that encourages re-examining the community's current emphasis on model complexity while highlighting the importance of training techniques.

\acknowledgments{This work is a result of the joint research project STADT:up (19A22006O). The project is supported by the German Federal Ministry for Economic Affairs and Energy (BMWE), based on a decision of the German Bundestag. The authors are solely responsible for the content of this publication.}

\clearpage


\bibliography{sections/mybib}  

\newpage
\appendix
\onecolumn

\begin{center}
    {\LARGE \bfseries Appendix}
\end{center}
\vspace{1.5em}

\section{Extended Related Work}
\label{app:extended_related_work}

This section provides further comparisons with related methods.
Table~\ref{tab:method_comparison} provides a structured comparison of DriveMLP against the most closely related methods, highlighting that while individual design choices appear in prior work, the specific combination of all components is novel to DriveMLP.

\begin{table}[ht]
\centering
\caption{Comparison of key design components across related methods. 
\textbf{CL Train} = closed-loop training; 
\textbf{Demo Sup.} = Supervision primarily provided by expert demonstrations (no traffic rules); 
\textbf{No Adv.} = no adversarial/iterative reward learning; 
\textbf{Lightweight} = $\leq$5M params, no generative or autoregressive architecture; 
\textbf{No Post-Proc.} = no rule-based post-processing at train or test time;
\textbf{Single Stage} = Single training stage;
\textbf{nuPlan} = evaluated on nuPlan or similar large-scale ego-planning benchmarks.
\checkmark\ = yes, \texttimes\ = no.
}
\label{tab:method_comparison}
\resizebox{\columnwidth}{!}{
\begin{tabular}{lccccccc}
\toprule
\textbf{Method} & CL Train & Demo Sup. & No Adv. & Lightweight & No Post-Proc. & Single Stage & nuPlan \\
\midrule
PDM-Open~\cite{Dauner2023CORL} / PlanTF~\cite{cheng2023plantf} / PLUTO~\cite{cheng2024pluto} w/o refine      & \texttimes & \checkmark & \checkmark & \checkmark & \checkmark & \checkmark & \checkmark \\
Diffusion~\cite{zheng2025diffusionbased} / Flow Planner~\cite{tan2025flow}  & \texttimes & \checkmark & \checkmark & \texttimes & \checkmark & \checkmark & \checkmark \\
AIRL~\cite{fu2018learning} / GAIL~\cite{GAIL2016Ermon}                      & \checkmark & \checkmark & \texttimes & \checkmark & \checkmark & \checkmark & \texttimes \\
CaRL~\cite{jaeger2025carl} / GIGAFLOW~\cite{cusumano-towner2025robust}      & \checkmark & \texttimes & \checkmark & \checkmark & \checkmark & \checkmark & \checkmark \\
BC-RL~\cite{imitationisnotenough}                                           & \checkmark & \checkmark & \checkmark & \checkmark & \checkmark & \checkmark & \checkmark \\
CarPlanner~\cite{carplanner}                                                & \checkmark & \checkmark & \checkmark & \texttimes & \texttimes & \checkmark & \checkmark \\
Peng et al.~\cite{peng2024improving}                                        & \checkmark & \checkmark & \checkmark & \texttimes & \checkmark & \texttimes & \texttimes \\
RITP~\cite{zeng2025reinforcedimitativetrajectoryplanning}                   & \checkmark & \checkmark & \checkmark & \checkmark & \texttimes & \texttimes & \checkmark \\
CIMRL~\cite{booher2024cimrl}                                                & \checkmark & \texttimes & \checkmark & \texttimes & \checkmark & \texttimes & \texttimes \\
CAT-K~\cite{zhang2025catk}                                                  & \texttimes & \checkmark & \checkmark & \texttimes & \checkmark & \texttimes & \checkmark \\
\midrule
\textbf{DriveMLP (ours)}                                                    & \checkmark & \checkmark & \checkmark & \checkmark & \checkmark & \checkmark & \checkmark \\
\bottomrule
\end{tabular}
}
\end{table}

\subsection{RL Benchmarks and Baselines for Autonomous Driving}

Several recent works have established RL baselines within AD simulation platforms.
Waymax~\cite{gulino2024waymax} provides a GPU-accelerated simulator built on the Waymo Open Dataset and includes DQN baselines with collision and off-road penalties; notably, these RL baselines underperform BC in their evaluation.
Scaling Is All You Need~\cite{harmel2023scaling} demonstrates JAX-accelerated RL for AD but requires BC pre-training and 2.5 billion agent steps, using multi-component handcrafted rewards.
V-Max~\cite{charraut2025vmax} proposes an RL framework with reward functions encoding traffic rules such as speed limits, lane-keeping, and comfort.
In all these cases, the supervision signal is fundamentally different from our single trajectory-distance reward, which does not encode any explicit traffic rules.

\subsection{Data Augmentation as an Alternative to Closed-Loop Training}

ChauffeurNet~\cite{bansal2019chauffeurnet} pioneered data augmentation via trajectory perturbations, a strategy adopted by recent methods including Diffusion Planner~\cite{zheng2025diffusionbased}.
While augmentation is substantially cheaper per sample (it does not require querying the current policy), it generates fixed, pre-defined perturbation patterns.
In contrast, RL exploration is adaptive to the current policy's weaknesses.
Our noise-robustness analysis (Section~\ref{sec:noise}) supports this: under significant action noise, DriveMLP maintains $>82$ Val14-R while Diffusion Planner, which uses 50\% augmented samples with perturbed-state recovery, drops to nearly 40.

\subsection{Supervised Fine-Tuning}

Zhang et al.~\cite{zhang2025catk} propose Closest-Among-Top-K (CAT-K), which selects the best trajectory from $K$ rollouts of a tokenized model.
This approach differs fundamentally from ours: CAT-K operates in a discrete action space of pre-computed trajectories, uses a two-stage training pipeline relying on imaginary rollouts within a learned model, and employs an autoregressive transformer with a GMM-parameterized policy.
Furthermore, their ego-policy variant is not conditioned on route information, making direct comparison on nuPlan non-trivial.

\subsection{Other Hybrid Approaches}

CIMRL~\cite{booher2024cimrl} combines IL and RL with safety constraints.
Their approach uses an IL pre-trained model to generate motion priors which are used to constrain RL exploration.
Similarly, RITP~\cite{zeng2025reinforcedimitativetrajectoryplanning} combines RL and IL by using a critic to score sampled candidate trajectories, training a transformer-based planner with imitation targets derived from the critic, and learning a Bayesian reward model with uncertainty penalization. 
It further employs a hybrid post-optimization stage with rule-based trajectory refinement at test time that boosts performance by more than 60\%. 

In contrast, we directly optimize a lightweight control policy with standard RL without requiring pre-trained models or rule-based refinement.

\subsection{Other Related Approaches}

DDPGfD~\cite{DBLP:journals/corr/VecerikHSWPPHRL17} and DAPG~\cite{Rajeswaran-RSS-18} use demonstrations to warm-start RL but still rely on hand-designed task rewards.
TD3+BC~\cite{fujimoto2021a} regularizes offline RL with imitation losses but cannot use closed-loop rollouts to address distribution shift.
SQIL~\cite{reddy2019sqil} uses binary imitation rewards ($r=1$ for expert data, $r=0$ otherwise), providing no gradient information about the degree of proximity to expert states, whereas our formulation provides dense, distance-sensitive feedback.
MaxEnt IRL~\cite{ziebart2008maximum}, Deep MaxEnt IRL~\cite{wulfmeier2015iros}, and ValueDICE~\cite{Kostrikov2020Imitation} require iterative reward learning or occupancy-matching objectives that remain computationally prohibitive at the scale of nuPlan.

\section{Theoretical discussion}
\label{app:theory}

This section provides a formal analysis of why closed-loop training with our trajectory-distance reward addresses key limitations of open-loop BC. 
We draw connections to DAgger~\cite{ross2011reduction}, inverse RL, and regularized BC, and analyze properties of the exponential reward formulation.

\subsection{Compounding Error in Open-Loop BC}

Standard BC minimizes a single-step loss over expert demonstrations:
\begin{equation}
    J_{\text{BC}}(\theta) = \mathbb{E}_{(s,a) \sim \mathcal{D}} \left[ \ell(\pi_\theta(s), a) \right],
    \label{eq:bc_objective}
\end{equation}
where $\mathcal{D}$ contains state-action pairs drawn from the expert's state distribution $d^{\pi^*}$.
At deployment, however, the policy $\pi_\theta$ induces its own distribution $d^{\pi_\theta}$, which diverges from $d^{\pi^*}$ due to imperfect actions.
Ross et al.~\cite{ross2011reduction} show that the resulting performance gap satisfies $J(\pi^*) - J(\pi_\theta) \leq 2 T^2 \epsilon_{\text{BC}}$, where $T$ is the planning horizon and $\epsilon_{\text{BC}}$ is the per-step error under $d^{\pi^*}$.
The quadratic dependence on $T$ is particularly severe for long planning horizons.

A further limitation is that the BC objective is entirely agnostic to the transition dynamics $\mathcal{T}$: actions are evaluated in isolation from their consequences.
This means the learned value function (implicit or explicit) may assign arbitrary values to off-distribution states, providing no corrective signal when the agent drifts~\cite{reddy2019sqil}.

\subsection{On-Policy Optimization Mitigates the Distribution Mismatch}

Our framework addresses the distribution shift problem by optimizing the policy through closed-loop rollouts:
\begin{equation}
    J_{\text{CL}}(\theta) = \mathbb{E}_{\tau \sim \pi_\theta} \left[ \sum_{t=0}^{T-1} \gamma^t \, r_t \right],
\end{equation}
where $\tau$ is generated by rolling out $\pi_\theta$ in the non-reactive simulator and $r_t$ is our trajectory-distance reward.
The policy gradient is computed under $d^{\pi_\theta}$, meaning the policy receives learning signal at exactly the states it encounters during deployment.
This reduces the distribution mismatch by construction, directly minimizing the performance gap rather than a surrogate loss under the wrong distribution.

\subsection{Connection to DAgger}

DAgger~\cite{ross2011reduction} addresses BC's distribution shift by collecting on-policy states from $\pi_\theta$ and querying the expert $\pi^*$ for corrective labels, achieving a linear bound $J(\pi^*) - J(\pi_\theta) \leq T \epsilon_{\text{DAgger}}$.
Our method achieves an analogous benefit \emph{without interactive expert access}.
The key insight is that in the non-reactive simulation setting, surrounding agents follow their recorded trajectories, so the logged expert ego trajectory remains a consistent reference.
The trajectory-distance reward $r_t = f(\|p_t^{\text{ego}} - p_t^{\text{expert}}\|_2)$ evaluates the policy's realized position against this fixed reference at every timestep, providing dense per-step supervision under $d^{\pi_\theta}$.

\subsection{Relationship to Inverse RL}

Adversarial IRL methods such as AIRL~\cite{fu2018learning} optimize a divergence between the policy's state occupancy measure $\rho^{\pi_\theta}$ and the expert's $\rho^{\pi^*}$, learning the reward via a discriminator.
Our trajectory-distance reward serves as an analytically defined surrogate for occupancy matching: it is maximized when $\rho^{\pi_\theta}$ concentrates on states near the expert trajectory.
In the limit where the reward sharpness $\alpha_1 \to \infty$, the positional reward $r_t^{\text{pos}} = \exp(-\alpha_1 \delta_t / (\kappa_t^{\text{expert}} + \epsilon))$ reduces to an indicator $\mathbb{1}[\delta_t = 0]$, recovering exact state matching.
For finite $\alpha_1$, the reward provides a smooth, dense signal that acts as a soft surrogate for occupancy matching.
This avoids the adversarial min-max optimization that makes GAIL and AIRL computationally expensive and unstable at scale.

Compared to SQIL~\cite{reddy2019sqil}, which assigns binary rewards ($r=1$ for expert states, $r=0$ otherwise), our formulation provides continuous gradient information proportional to the deviation magnitude.
From a policy-gradient perspective, dense per-timestep rewards distribute learning signal across all rollout steps, reducing gradient variance compared to the sparse signal of SQIL, where most updates carry no reward information~\cite{greensmith2004variance}.

\subsection{Interpretation as Dynamics-Aware Regularized BC}

Following the framework of Reddy et al.~\cite{reddy2019sqil} and Piot et al.~\cite{piot2014boosted}, who showed that RL with imitation rewards is equivalent to BC regularized by a Bellman-consistency prior, we observe that our closed-loop objective can also be decomposed into two functional components:
(1)~an \emph{imitation term}, where the trajectory-distance reward encourages actions that keep the policy close to the expert; and 
(2)~a \emph{dynamics regularization term}, where multi-step rollouts enforce that actions are evaluated by their consequences under the true transition dynamics $\mathcal{T}$.
The second component is entirely absent in BC, where action quality is assessed independently at each timestep.

\subsection{Properties of the Exponential Reward}
The exponential formulation $r_t^{\text{pos}} = \exp(-\alpha_1 \delta_t / (\kappa_t^{\text{expert}} + \epsilon))$ provides a more suitable reward landscape for trajectory tracking compared to the inverse distance $\tilde{r}_t = \kappa_t^{\text{expert}}/(\delta_t + \epsilon)$ or the linear penalty $\tilde{r}_t = -\delta_t$ used by Peng et al.~\cite{peng2024improving}.

In driving scenarios with multiple valid trajectory modes (e.g., passing an obstacle on either side), a policy that averages between modes produces trajectories that follow neither.
For two modes separated by distance $2d$, the reward ratio between committing to one mode and averaging between them is $\exp(\alpha_1 \cdot d / (\kappa_t^{\text{expert}} + \epsilon))$ under our formulation.
This ratio grows exponentially with mode separation, creating a strong incentive for the policy to commit to a single consistent trajectory.
Under the inverse distance reward, the same ratio grows linearly, providing a weaker preference for commitment, while the linear reward $-\delta_t$, which is additionally unnormalized, leads to unstable optimization in practice.

Our empirical ablation supports this analysis: the exponential reward outperforms the inverse distance by $+5.5$ NR / $+3.8$ R points and the unnormalized linear penalty by $+39.9$ NR / $+38.9$ R points on nuPlan Val14 (Table~\ref{tab:reward}).

\section{Experimental Setup}
\label{app:setup}

\subsection{Implementation Details}

To enable scalable distributed RL training~\cite{48990} on nuPlan we employ the Ray RLlib~\cite{liang2021rllib} implementation of PPO. 
For Highway-Env, we use SAC~\cite{stable-baselines3} to demonstrate the generality of our approach across different RL algorithms.
BC and AIRL baselines are implemented using the Imitation library~\cite{gleave2022imitation}, with the same model architectures as our method.

During training, the policy is executed at 1~Hz by repeating each selected action for 10 consecutive simulation steps.
This action repetition promotes greater diversity of state transitions within each training batch (see Section~\ref{app:param_sensitivity} for an ablation).
Evaluation is carried out at 10~Hz strictly following nuPlan settings.

The ego transition model follows nuPlan's default kinematic bicycle model, using a Numba-accelerated implementation during training.

PPO improves upon strict on-policy methods (e.g., REINFORCE~\cite{Williams1992} and A2C~\cite{pmlrAsynchronous}) by enabling multiple gradient updates per iteration while using clipping to control policy divergence. 
We perform approximately 2,000 updates per iteration, substantially improving sample reuse and data efficiency.

We train our approach for approximately 100K environment steps on Highway-Env and for 100M environment steps on nuPlan. 
We set the discount factor $\gamma = 0.7$, which corresponds to a planning horizon of approximately 3.3 seconds, similar to the horizon used by other planners.

\subsection{Evaluation Metrics}

For \textbf{Highway-Env}, we report:
(1)~mean cumulative reward, 
(2)~mean episode length, and 
(3)~termination rate due to collisions or off-road events. 
All methods are evaluated over 100 episodes in the reactive simulator, with results averaged across three random seeds.

For \textbf{nuPlan}, we follow the official closed-loop evaluation protocol~\cite{nuplan} and report the overall score under both reactive and non-reactive settings for three standard splits: Val14~\cite{Dauner2023CORL}, Test14-Hard~\cite{cheng2023plantf}, and Test14~\cite{cheng2023plantf}.

\subsection{Hyperparameters}
\label{app:hyperparams}

Table~\ref{tab:hparams} reports the main training hyperparameters used in the experiments.
Hyperparameters not reported here use the default value of the library implementation.

\begin{table}[ht]
\centering
\caption{Hyperparameters used in Nuplan and Highway env experiments}
\label{tab:hparams}
\resizebox{\columnwidth}{!}{
\begin{tabular}{lcccccccc}
\toprule
& \multicolumn{4}{c}{Highway-Env} & \multicolumn{4}{c}{nuPlan}\\
\cmidrule(lr){2-5} \cmidrule(lr){6-9}
Name & BC & AIRL & BC-RL & DriveMLP & BC & AIRL & BC-RL & DriveMLP \\
\midrule
Algorithm & BC & AIRL (SAC) & BC+SAC & SAC & BC & AIRL (PPO) & BC+PPO & PPO \\
Training steps & - & 300K & 100K & 100K & - & 100M & 100M & 100M\\
Learning rate & 0.001 & 0.0003 & 0.0003 & 0.0003 & 0.0001 & 0.00005 & 0.00005 & 0.00005\\
Learning rate schedule & exponential decay & - & - & - & exponential decay & cosine decay & cosine decay & cosine decay\\
Rollout buffer size & - & 1M & 1M & 1M & - & 98K & 98K & 98K\\
Batch size & 64 & 256 & 256 & 256 & 512 & 512 & 512 & 512\\
Epochs & 80 & - & - & - & 100 & 10 & 10 & 10\\
Discount factor & - & 0.7 & 0.7 & 0.7 & - & 0.7 & 0.7 & 0.7\\
Gae lambda & - & - & - & - & - & 0.5 & 0.5 & 0.5\\
BC/RL update ratio & - & - & 8 & - & - & - & 8 & -\\
Discriminator updates & - & 4 &- & - & - & 1 & - & -\\
Discriminator learning rate & - & 1e-2 &- &- &- & 3e-5 & - & -\\

\bottomrule
\end{tabular}
}
\end{table}

Unlike traditional rule-based rewards in AD which often involve conflicting objectives (e.g., progress vs. safety), our trajectory-based reward terms are inherently aligned. 
Both $r_t^{pos}$ and $r_t^{vel}$ are maximized at the expert's state, meaning they share the same global optimum. 
Consequently, the hyperparameters $\alpha_1, \alpha_2,$ and $\alpha_3$ function as scaling factors that regulate the smoothness of the reward surface rather than shifting the task's objective.
We used the same values across all our experiments: $\alpha_1=1/3, \alpha_2=1/4, \alpha_3=0.7$.
We verified that the proposed reward is not sensitive to scaling: setting $\alpha_1{=}\alpha_2{=}1$,  and removing $\alpha_3$ yields a Val14-R score of 85.87.
The terminal penalty value is $r_p=-1$.
The deviation-based terminal condition is triggered when ego velocity is not within a 100\% tolerance margin of the expert velocity, namely when $|v_t^{\text{ego}}/v_t^{\text{expert}} - 1|>1.0$ for $v_t^{\text{expert}}>1 m/s$.

\subsection{Datasets}
\label{app:dataset}

For \textbf{Highway-env}, we construct an expert demonstration set by rolling out the built-in IDM policy in the \emph{reactive} simulator for 1000 episodes of 30s each. 
We discard episodes that terminate due to collisions or off-road events, yielding approximately 750 optimal trajectories used for training.

For \textbf{nuPlan}, we follow standard preprocessing protocols to enable direct comparison with prior work. 
For open-loop BC, we use the same scenario extraction procedure as Diffusion Planner~\cite{zheng2025diffusionbased}, resulting in 1M training scenarios. 
For closed-loop training, we adopt the CaRL preprocessing setup~\cite{jaeger2025carl}: we select all temporally non-overlapping scenarios of 15s duration from the nuPlan training split that contain valid route information, resulting in approximately 270K scenarios.

\section{InterPlan Results}
\label{app:interplan}
We additionally evaluate on InterPlan~\cite{10803052}, which augments nuPlan scenarios with challenging long-tail interactions (Table~\ref{tab:interplan}). 
DriveMLP exceeds the reported PlanTF and Diffusion Planner scores on both splits. 
On the official 80-scenario split, its mean score is close to Flow Planner's but below PLUTO's; on the full 335-scenario set, it is close to PLUTO's and below Flow Planner's. 
These results provide additional evidence of transfer to challenging interactive scenarios.

\begin{table}[h]
\centering
\caption{Interplan benchmark results. $^\dagger$PLUTO incorporates centerline reference paths as an additional input.}
\label{tab:interplan}
\resizebox{0.7\columnwidth}{!}{
\begin{tabular}{lccccc}
\toprule
Split & PlanTF & PLUTO w/o ref.$^\dagger$ & Diff. Planner & Flow Planner & \textbf{DriveMLP} \\
\midrule
80 scenarios & 30.53 & 49.27 & 24.71 & 36.59 & 37.07{\tiny$\pm$2.36} \\
335 scenarios & 47.70 & 58.47 & 52.90 & 61.82 & 58.43{\tiny$\pm$2.74} \\
\bottomrule
\end{tabular}
}
\end{table}

\section{Model Capacity Analysis}
\label{app:agnosticism}

This section complements the backbone comparisons in Section~\ref{sec:ctrl_comp2} by examining MLP capacity scaling and a reduced transformer policy.

\subsection{MLP Capacity Scaling}
To examine the effects of model capacity and training strategy, we vary the capacity of the MLP architecture used by DriveMLP from 112K to 8.7M parameters, comparing EGRL with BC at each capacity.
Table~\ref{tab:mlp_capacity} reports reactive performance, improvements over BC, inference latency, and training cost.

\begin{table}[h]
\centering
\caption{MLP capacity scaling under EGRL. Improvements over BC are reported in score points.}
\label{tab:mlp_capacity}
\resizebox{0.6\columnwidth}{!}{
\begin{tabular}{lccccccc}
\toprule
Size & 112K & 247K & 590K & 1.1M & \textbf{2.3M} & 4.6M & 8.7M\\
\midrule
Val14-R         & 82.22 & 84.39 & 84.96 & 85.74 & \textbf{86.51} & 86.18 & 86.33 \\
$\Delta$ vs. BC & +53.31 & +54.54 & +53.20 & +53.70 & \textbf{+54.15} & +53.85 & +54.22 \\
Infer. (ms)     & 1.86 & 1.95 & 2.27 & 2.39 & \textbf{2.76} & 2.99 & 3.47 \\
Train (L4-h)    & 56 & 60 & 74 & 87 & \textbf{112} & 155 & 185 \\
\bottomrule
\end{tabular}
}
\end{table}

Increasing capacity under BC yields marginal gains, whereas EGRL improves reactive performance over BC by approximately 53--55 points across the tested capacities.
EGRL performance increases up to the 2.3M-parameter model, while larger models provide no further improvement in the observed score despite higher inference latency and training cost.

\subsection{Reduced Transformer Policy}

Starting from the Diffusion Planner encoder~\cite{zheng2025diffusionbased}, we reduce the policy capacity to approximately 500K parameters to obtain inference latency comparable to DriveMLP.
We use the encoder and control-head adaptations described in Section~\ref{sec:ctrl_comp2}, without the diffusion generation process or auxiliary multi-agent prediction objective.
As shown in Table~\ref{tab:nuplan_transformer}, EGRL achieves approximately +21/+47 (NR/R) points over the BC baseline.
This provides additional evidence that the benefits extend beyond MLP policies, including a reduced transformer architecture.

\begin{table}[h]
\centering
\caption{Reduced transformer policy (500K parameters) on nuPlan Val14. NR = Non-Reactive, R = Reactive simulation.}
\label{tab:nuplan_transformer}
\resizebox{0.25\columnwidth}{!}{
\begin{tabular}{lcc}
\toprule
Method & NR & R \\
\midrule
BC & 61.25 & 32.44\\
\textbf{EGRL} & \textbf{82.01} & \textbf{79.68} \\
\bottomrule
\end{tabular}
}
\end{table}

\section{Ablations and Parameter Analysis}
\label{app:param_sensitivity}

\subsection{Reward Components and Hyperparameter Sensitivity}

Table~\ref{tab:ablations} summarizes the effects of removing individual episode termination conditions.
These conditions improve performance while restricting rollouts to safe and approximately log-consistent states.
Collision termination prevents rollouts from continuing through physically invalid vehicle overlaps, while off-road termination restricts exploration to the drivable area.
Deviation termination limits exploration in states where interactions with surrounding agents, which continue to replay their logged trajectories independently of the ego, may become inconsistent with the recorded scenario.
This differs from constructing a dense reward by weighting traffic-rule objectives such as speed-limit compliance, comfort, lane keeping, and progress.
Such objectives often require context-dependent trade-offs.
For example, yielding to an emergency vehicle may require leaving the current lane or, where legally permitted, proceeding past a red light; comfort may also need to be subordinated to the maneuver.
Encoding these exceptions can require scenario-specific reward engineering and weighting.
Our approach instead derives dense behavioral supervision from expert trajectories, while retaining boolean termination conditions to constrain training rollouts.

\begin{table}[ht]
\centering
\caption{Episode termination ablations for DriveMLP on nuPlan Val14.}
\label{tab:ablations}
\resizebox{0.3\textwidth}{!}{%
\begin{tabular}{lcc}
\toprule
Component & NR & R \\
\midrule
\textbf{Base}  &  \textbf{90.38} & \textbf{86.51} \\
\midrule
w/o off-road       & 80.76 & 78.67 \\
w/o collision      & 80.08 & 79.14 \\
w/o deviation      & 86.92 & 84.11 \\
\bottomrule
\end{tabular}%
}
\end{table}

In Table \ref{tab:gamma} we analyze the sensitivity to the discount factor and in Table \ref{tab:penalty} to the terminal penalty.
As reported by~\cite{jaeger2025carl}, AD rewards exhibit significant sensitivity to mini-batch size variations with benchmark scores dropping near zero. 
Table~\ref{tab:batch} shows that our trajectory-based reward formulation is substantially more robust to this hyperparameter.

\begin{table}[ht]
\centering
\begin{minipage}{0.3\columnwidth}
\centering
\caption{nuPlan Val14 score at varying magnitude of the discount factor.}
\label{tab:gamma}
\resizebox{\columnwidth}{!}{
\begin{tabular}{lc}
\toprule
Discount factor & R Score \\
\midrule
$\gamma=0.6$    & 85.08  \\
\textbf{$\gamma=0.7$}   &  \textbf{86.51}  \\
$\gamma=0.8$  & 84.48   \\
\bottomrule
\end{tabular}
}
\end{minipage}
\hspace{10pt}
\begin{minipage}{0.3\columnwidth}
\centering
\caption{nuPlan Val14 score at varying magnitude of the terminal penalty.}
\label{tab:penalty}
\resizebox{\columnwidth}{!}{
\begin{tabular}{lc}
\toprule
Terminal penalty &  R Score \\
\midrule
$r_p=-0.5$    & 85.57 \\
\textbf{$r_p=-1$}    & \textbf{86.51}  \\
$r_p=-2$   & 84.39   \\
\bottomrule
\end{tabular}
}
\end{minipage}
\hspace{10pt}
\begin{minipage}{0.3\columnwidth}
\centering
\caption{nuPlan Val14 score at varying mini-batch size.\\}
\label{tab:batch}
\resizebox{\columnwidth}{!}{
\begin{tabular}{lcc}
\toprule
Mini-batch size & R Score \\
\midrule
128    & 84.02 \\
512    & \textbf{86.51}  \\
2048   & 82.49   \\
\bottomrule
\end{tabular}
}
\end{minipage}
\end{table}

\subsection{Action Repetition Ablation}
\label{app:repetition}

We train the policy using the nuPlan simulator at its native 10\,Hz control frequency, but execute actions at a reduced rate of 1\,Hz by repeating each selected action for 10 consecutive simulation steps.
At evaluation time, we use the standard nuPlan setup and run the policy at 10\,Hz without action repetition (i.e., each action is applied once per simulator step).

This action repetition introduces a trade-off between spatial coverage and control fidelity.
Lower control frequency increases the displacement between consecutive decisions, broadening the state distribution within each training batch.
It also reduces the effective RL horizon (e.g., 15 steps for a 15-second scenario), which simplifies credit assignment and improves training stability, analogous to frame-skipping in Atari~\cite{mnih2015human}.
Conversely, higher control frequencies more closely match the 10\,Hz evaluation setting.
Table~\ref{tab:repetition} compares performance at 0.5, 1, and 2\,Hz, showing that 1\,Hz provides the best balance.

\begin{table}[ht]
\centering
\caption{Action frequency ablation: nuPlan Val14.}
\label{tab:repetition}
\resizebox{0.3\columnwidth}{!}{
\begin{tabular}{lcc}
\toprule
Frequency & NR & R \\
\midrule
0.5 Hz   & 75.23 &  77.61 \\
\textbf{1 Hz}   &  \textbf{90.38} & \textbf{86.51} \\
2 Hz  & 85.13  &  82.28 \\
\bottomrule
\end{tabular}
}
\end{table}

While our uniform 1\,Hz repetition strategy proves highly effective, a promising future direction is to dynamically parameterize this temporal abstraction using frameworks like FiGAR~\cite{sharma2017learning}. 
Incorporating variable-rate repetition could allow the policy to adaptively scale its control frequency, deploying high-frequency adjustments during critical maneuvers while maintaining a sparse, sample-efficient temporal horizon in nominal driving scenarios.

\subsection{Ego-state Ablation}
\label{app:ego_state}
Table~\ref{tab:ego_state} reports an ego-state ablation.
Ego-state information improves DriveMLP by $+11$ NR points, but hurts open-loop trained planners: both the BC MLP and BC Transformer degrade when ego-state features are included, consistent with Diffusion Planner's reported performance drop of more than 10 points~\cite{zheng2025diffusionbased}.
This is consistent with the hypothesis that closed-loop RL training better captures the interplay between ego dynamics and planning decisions, whereas open-loop BC may learn spurious correlations from ego-state features.

\begin{table}[ht]
\centering
\caption{Ego-state ablation: nuPlan Val14 NR.}
\label{tab:ego_state}
\resizebox{0.65\columnwidth}{!}{
\begin{tabular}{lccc}
\toprule
Architecture & Training Method & w/o Ego State & w/ Ego State \\
\midrule
DriveMLP & BC & 54.90 &  53.64 \\
Transformer & BC &  61.25 & 54.76 \\
DriveMLP & EGRL & 78.79  &  90.38 \\
\bottomrule
\end{tabular}
}
\end{table}

\section{Noise Robustness Visualization}
\label{app:noise}
\begin{figure}[hbt]
  \begin{center}
    \centerline{\includegraphics[width=\columnwidth]{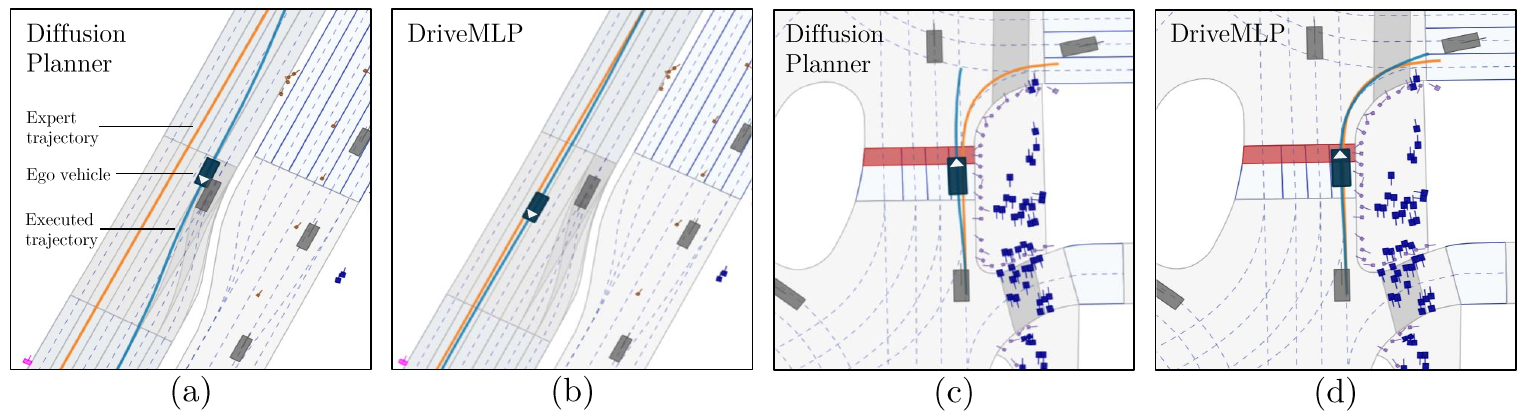}}
    \caption{
       Noise brings the Diffusion Planner close to a neighboring agent, with no corrective response to avoid the collision (a), whereas DriveMLP remains close to the expert trajectory (b).
       Similarly, noise can cause the Diffusion Planner to miss a turn (c), while our planner remains largely unaffected (d). 
    }
    \label{fig:noise_nuboard}
  \end{center}
  \vskip -0.2in
\end{figure}

Figure \ref{fig:noise_nuboard} provides a qualitative comparison of representative trajectories. 
Despite the injected noise, our method produces trajectories that remain closely aligned with the expert, whereas Diffusion Planner exhibits a gradual drift away from the reference trajectory. 

\section{Reproducibility and Simulation Infrastructure}
\label{app:reproducibility}

\textbf{Hardware and Software.} We used Python 3.9 together with PyTorch 2.7.1 and CUDA 12.6 for GPU acceleration.
All nuPlan experiments were run on a single compute node with 4 NVIDIA L4 GPUs and 48 CPU cores.

\textbf{Training Parallelism and Runtime.}
For the nuPlan experiments, we parallelize rollouts across 45 environments using Ray's asynchronous vectorization interface, which is well suited to non-homogeneous environments~\cite{Wijmans2020DD-PPO}.
Training runs for 100M environment steps and completes in approximately 30 hours end-to-end; with our current setup, PPO spends less than half its wall-clock time on data collection.
This is made feasible by a domain-specific simulation runtime that achieves more than 100$\times$ higher throughput than a naive nuPlan wrapper with vectorized observation features (e.g., CaRL nuPlan Gym Env. with Diffusion Planner observation builder), as shown in Table~\ref{tab:speedup}.
As recent progress in GPU-accelerated simulation has shown~\cite{kazemkhani2025gpudrive}, and through asynchronous policy optimization such as APPO~\cite{Luo2020IMPACT}, there is significant headroom for further speedup.

\begin{table}[ht]
\centering
\caption{Single env. reset/step functions performance}
\label{tab:speedup}
\resizebox{0.4\columnwidth}{!}{
\begin{tabular}{lcc}
\toprule
Env & Reset & Step \\
\midrule
Naive stack (ms) & 559.4 &  366.27  \\
\textbf{Ours} (ms) & \textbf{3.9} & \textbf{1.59} \\
\midrule
Speedup & 143x & 230x \\
\bottomrule
\end{tabular}
}
\end{table}

\textbf{Simulation Optimizations.}
The speedup comes from five main engineering changes.
\textbf{(1)} We replace the generic nuPlan/CaRL orchestration stack with a domain-specific runtime optimized for low-latency stepping. 
This removes unnecessary planner-state reconstruction, bypasses general-purpose map interfaces, and reduces the reset and step overhead.
\textbf{(2)} Expensive scenario parsing and map preprocessing are moved offline. 
During training, the simulator operates on preconverted scenarios and cached map artifacts, avoiding repeated reconstruction of runtime state.
\textbf{(3)} Pointer-heavy Python object structures are replaced with contiguous NumPy buffers, which improve memory locality and reduce allocation overhead.
\textbf{(4)} Online retrieval operations use spatial indexing and candidate pruning instead of full-set sorting. 
Broad-phase culling is applied before exact geometric checks.
\textbf{(5)} Latency-critical components, including lane ranking, coordinate transforms, vehicle propagation, collision checks, and off-road detection, are implemented using batched linear algebra and Numba-JIT kernels~\cite{numba}.

Despite these optimizations, the entire simulation stack remains implemented in pure Python, without C/C++ extensions, preserving ease of modification and integration into existing research workflows.

\textbf{Inference Benchmarking Protocol.} Inference latency in Table~\ref{tab:inference} is measured on an Intel Xeon 6136 CPU and an NVIDIA Quadro P6000 GPU.
All methods are evaluated with the same batch size and timing procedure under identical conditions.

\section{Failure Cases}
\label{app:failure}

\begin{figure}[hbt]
  \begin{center}
    \centerline{\includegraphics[width=\columnwidth]{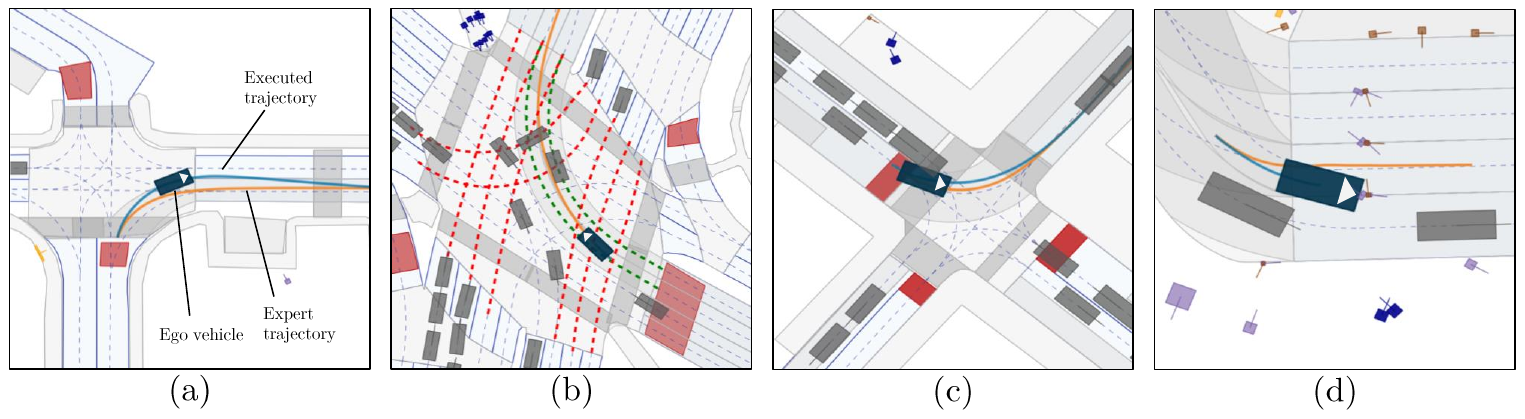}}
    \caption{
    Representative failure cases of DriveMLP. 
      Expert trajectory is shown in orange, while the trajectory executed by ego is in blue. 
      Failure modes include (a) overshooting turns and invading the opposite lane, (b) failure to progress due to a limited time window in which the intersection is clear, (c) side collisions with other vehicles, and (d) late braking for obstacles in the lane. 
      All scenarios are taken from the Val14-Reactive evaluation split.
    }
    \label{fig:failures}
  \end{center}
  \vskip -0.2in
\end{figure}

Despite the overall improvements, challenges remain.
Figure~\ref{fig:failures} illustrates representative failure scenarios that reveal current limitations and opportunities for future development.
The observed failure modes include overshooting turns and invading the opposite lane, failure to progress through narrow time windows at intersections, side collisions with other vehicles, and late braking for obstacles.

Table~\ref{tab:detailed_metrics} provides a per-metric and per-scenario breakdown comparing DriveMLP against Diffusion Planner~\cite{zheng2025diffusionbased} (reproduced) on Val14 and Test14-Hard.
DriveMLP surpasses Diffusion Planner on \textit{TTC} and \textit{Drivable Area Compliance} across all splits, but lags slightly behind on \textit{Driving Direction Compliance}.

The \textit{Starting Left Turn} scenario is not present in the nuPlan training distribution, making it a natural test of out-of-distribution generalization.
DriveMLP matches Diffusion Planner's almost exactly: 88.55/74.08 (NR/R) vs.\ 86.71/72.49 on Val14, and 65.36/72.21 vs.\ 65.73/67.01 on Test14-Hard. This parity indicates that model complexity alone does not determine generalization to unseen maneuvers: a lightweight policy trained in closed-loop can achieve similar out-of-distribution performance to a significantly more complex generative model.

On Val14-Reactive, DriveMLP also exceeds the score obtained by replaying the logged expert trajectory (86.51 vs.\ 82.02).

\begin{table}[ht]
\centering
\caption{Performance Comparison: DriveMLP vs. Diffusion Planner across nuPlan Splits}
\label{tab:detailed_metrics}
\resizebox{\textwidth}{!}{%
\begin{tabular}{l|cc|cc||cc|cc}
\toprule
 & \multicolumn{4}{c||}{\textbf{DriveMLP}} & \multicolumn{4}{c}{\textbf{Diffusion Planner}} \\
 & \multicolumn{2}{c|}{\textbf{Val14}} & \multicolumn{2}{c||}{\textbf{Test-Hard}} & \multicolumn{2}{c|}{\textbf{Val14}} & \multicolumn{2}{c}{\textbf{Test-Hard}} \\
 \textbf{Metrics} & \textbf{NR} & \textbf{R} & \textbf{NR} & \textbf{R} & \textbf{NR} & \textbf{R} & \textbf{NR} & \textbf{R} \\
\midrule
\textbf{Overall Score} & 90.39 & 86.51 & 76.69 & 74.51 & 89.86 & 82.95 & 75.73 & 68.94 \\
Drivable Area Compliance & 99.82 & 99.82 & 97.43 & 97.79 & 97.94 & 97.85 & 95.59 & 94.85 \\
Driving Direction Comp. & 99.37 & 99.42 & 97.79 & 97.98 & 99.51 & 99.82 & 98.90 & 98.90 \\
Ego is Comfortable & 87.39 & 88.10 & 86.03 & 87.13 & 92.49 & 89.36 & 91.91 & 86.40 \\
Ego is Making Progress & 98.93 & 98.57 & 95.22 & 94.49 & 99.91 & 96.69 & 98.53 & 92.28 \\
Ego Progress vs Expert & 91.27 & 88.02 & 82.40 & 77.19 & 95.29 & 85.98 & 89.77 & 76.60 \\
No At-Fault Collisions & 97.27 & 94.45 & 92.10 & 90.26 & 96.20 & 93.11 & 87.68 & 86.21 \\
Speed Limit Compliance & 97.75 & 98.28 & 97.03 & 97.18 & 96.52 & 98.29 & 96.44 & 97.39 \\
TTC Within Bound & 94.28 & 89.71 & 83.09 & 83.82 & 91.59 & 88.01 & 77.57 & 80.51 \\
\midrule
\textbf{Scenario Scores} & & & & & & & & \\
Behind Long Vehicle & 99.17 & 99.60 & 90.63 & 91.14 & 99.13 & 98.58 & 91.08 & 86.42 \\
Changing Lane & 89.07 & 88.80 & 73.27 & 73.59 & 85.64 & 84.57 & 64.79 & 76.63 \\
Following Lane w/ Lead & 97.33 & 88.74 & 88.84 & 89.42 & 95.65 & 80.70 & 90.36 & 79.69 \\
High Lateral Accel. & 84.39 & 83.90 & 59.64 & 58.25 & 85.71 & 83.26 & 58.44 & 51.05 \\
High Magnitude Speed & 96.13 & 93.40 & 91.31 & 84.62 & 92.84 & 90.54 & 88.18 & 78.94 \\
Low Magnitude Speed & 88.55 & 88.67 & 47.25 & 54.91 & 91.36 & 82.98 & 50.75 & 48.01 \\
Near Multiple Vehicles & 95.01 & 83.69 & 77.52 & 60.06 & 95.47 & 83.33 & 66.74 & 50.52 \\
Starting Left Turn & 88.55 & 74.08 & 65.36 & 72.21 & 86.71 & 72.49 & 65.73 & 67.01 \\
Starting Right Turn & 83.40 & 76.38 & 69.79 & 57.30 & 84.07 & 68.61 & 67.27 & 56.11 \\
Intersection Traversal & 93.98 & 87.66 & 83.93 & 80.64 & 93.65 & 82.15 & 88.56 & 89.07 \\
Stationary in Traffic & 97.54 & 94.72 & 96.47 & 83.11 & 98.17 & 93.42 & 97.66 & 71.52 \\
Stopping w/ Lead & 96.11 & 98.47 & 98.19 & 98.12 & 95.34 & 94.61 & 99.07 & 97.61 \\
Traversing Pickup/Dropoff & 78.94 & 80.00 & 67.38 & 68.96 & 76.29 & 75.76 & 59.41 & 70.03 \\
Waiting for Pedestrian & 91.88 & 87.92 & 69.20 & 76.97 & 91.48 & 83.30 & 78.14 & 47.86 \\
\bottomrule
\end{tabular}%
}
\end{table}

\end{document}